\documentclass[lettersize,journal]{IEEEtran}
\usepackage{amsmath,amsfonts}
\usepackage{algorithmic}
\usepackage{algorithm}
\usepackage{array}
\usepackage{textcomp}
\usepackage{stfloats}
\usepackage{url}
\usepackage{verbatim}
\usepackage{graphicx}
\usepackage{cite}
\usepackage{placeins}
\usepackage{array}
\usepackage{graphicx}
\usepackage{multirow}
\usepackage{caption}

\usepackage{tikz}
\usetikzlibrary{shapes.geometric, arrows}

\newcommand\MyBox[2]{
  \fbox{\lower0.75cm
    \vbox to 1.7cm{\vfil
      \hbox to 1.7cm{\hfil\parbox{1.4cm}{#1\\#2}\hfil}
      \vfil}%
  }%
}

\tikzstyle{startstop} = [rectangle, rounded corners, minimum width=3cm, minimum height=0.8cm, text centered, draw=black, fill=orange!0]
\tikzstyle{process} = [rectangle, minimum width=3cm, minimum height=0.7cm, text centered, draw=black, fill=blue!0]
\tikzstyle{arrow} = [thick, ->, >=stealth]
\usepackage{float}
\begin{document}

\title{Integrating Large Language Models into a Tri-Modal Architecture for Automated Depression Classification}

\author{Santosh V. Patapati

}

\markboth{}%
{Shell \MakeLowercase{\textit{et al.}}: A Novel LLM-Based Multi-Modal Architecture for Depression Diagnosis}

\maketitle

\begin{abstract}
Major Depressive Disorder (MDD) is a pervasive mental health condition that affects 300 million people worldwide. This work presents a novel, BiLSTM-based tri-modal model-level fusion architecture for the binary classification of depression from clinical interview recordings. The proposed architecture incorporates Mel Frequency Cepstral Coefficients, Facial Action Units, and uses a two-shot learning based GPT-4 model to process text data. This is the first work to incorporate large language models into a multi-modal architecture for this task. It achieves impressive results on the DAIC-WOZ AVEC 2016 Challenge train/validation/test split and Leave-One-Subject-Out cross-validation split, surpassing all baseline models and multiple state-of-the-art models. In Leave-One-Subject-Out testing, it achieves an accuracy of 91.01\%, an F1-Score of 85.95\%, a precision of 80\%, and a recall of 92.86\%.
\end{abstract}


\section{Introduction}
\IEEEPARstart{M}{ajor} Depressive Disorder (MDD) is a pervasive mental health condition that affects 300 million people worldwide \cite{Chodavadia2023}. The COVID-19 pandemic has further exacerbated this issue, leading to a staggering 27\% increase in the global prevalence of MDD \cite{Santomauro2021}.

Unlike biological disorders, MDD is not diagnosed through blood tests or imaging. It lacks these gold standards. Instead, the most common approach for MDD diagnosis relies on clinical interviews \cite{Guidi2011} and self-reported inventories (SRIs) \cite{Radloff1977, Kroenke2001}. These systems have been subject to criticism for their subjectivity \cite{Allsopp2019, De_Silva2017, Fuchs2010}, which gives way for a number of issues. Bias, for example, can come from both parties. In clinical interviews and SRIs, patients may exaggerate or under-report symptoms in social desirability bias \cite{Althubaiti2016, Latkin2017}. Confirmation bias affects how clinicians weigh a patient's symptoms against the assigned criteria \cite{Mendel2011}. As a result, the misdiagnosis rate of MDD is estimated to be as high as 54\% \cite{Ayano2021}.

In recent years, there has been a growing interest in the use of machine learning (ML) systems to automatically assess the presence and severity of MDD. This offers a low-cost and objective alternative to current methods. A heavily researched use case of ML involves making diagnoses from clinical interview recordings.

Numerous studies investigate a multi-modal approach, which combines both verbal and non-verbal cues to reach a diagnosis. These models achieve extremely high levels of success. Currently, the majority of state-of-the-art models take in three modes of input: Audio, video, and text-based data.

The text-based modality is generally seen as the weakest link in diagnosing MDD from clinical interviews \cite{Muzammel2021}. This can be attributed to the scarcity of task-specific text-based training data. This lack of data hinders the effective training of NLP models, which are data-intensive, resulting in suboptimal performance compared to the audio and video modalities.

To date, no research has attempted to integrate Large Language Models (LLMs) into a multi-modal architecture for this task. Due to their training on large corpora, we hypothesized incorporating LLMs into such an architecture would improve the accuracy of depression diagnosis from clinical interview recordings and alleviate the issue of scarce training data.

In this work, we make the following contributions:
\begin{enumerate}
    \item We propose a novel, tri-modal architecture that utilizes LLMs. This is the first work to incorporate LLMs into a multi-modal framework for this task.
    \item We demonstrate that LLMs are effective for mental health diagnosis when incorporated into multi-modal architectures. The proposed architecture outperforms baseline and state-of-the-art models tested on the AVEC 2016 Challenge dataset and Leave-One-Subject-Out Cross-Validation. The results of this work serve as an indicator for the potential of such an architecture in classifying depressed patients.
    \item The proposed architecture is integrated into a locally hosted web application to emulate the potential use of such a model in the real-world.
\end{enumerate}

\section{Related Works}

\begin{figure*}[!t]
\centering
\includegraphics[width=5.5in]{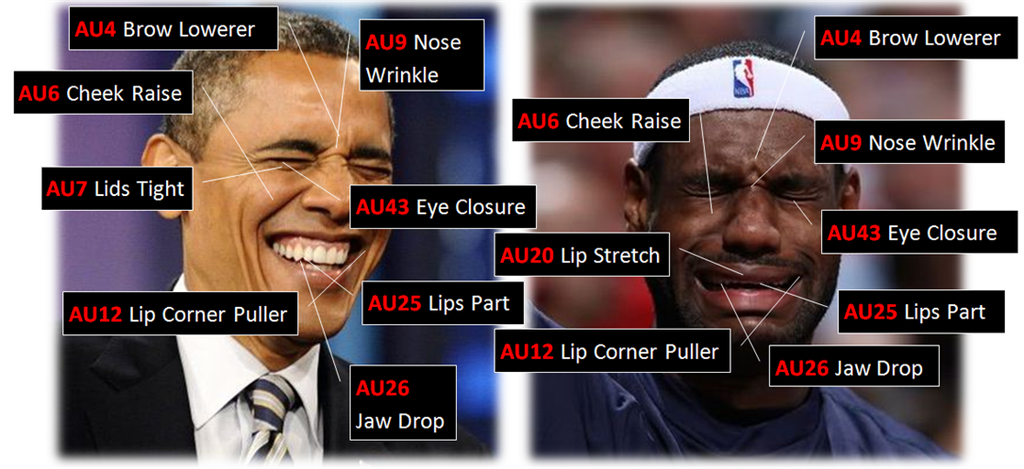}

\caption{Example of Facial Action Unit descriptors in famous photos \cite{Tu2019}. They can code nearly any anatomically possible facial expression.}
\label{fig1}
\end{figure*}

\begin{table*}[!ht]
\caption{State-of-the-art multi-modal fusion systems for binary depression classification. (*Classification done with Leave-One-Subject-Out Cross-Validation}
\centering
\begin{tabular}{|c|c|c|c|c|c|}
\hline
Paper & Architecture & Fusion Level & Reported Performance & Modalities & Features\\
\hline
\hline
(Gimeno-Gómez et al.) \cite{Gimeno} & Transformer & Late & F1-Score: 72& Audio & COVAREP \cite{Degottex2014} \\
 & & & Precision: 72&  & Vocal Tract Resonance Frequencies \cite{Sataloff2017} \\
 & & & Recall: 72& Video & 3D Facial Landmarks \\
 & & & & & FAUs \\
 & & & & & Gaze Tracking \\
 & & & & & Head Pose Landmarks \\
\hline
\hline
(Muzammel et al.) \cite{Muzammel2021} & LSTM & Model & Accuracy: 77.16& Audio & MFCCs\\
 &  &  &  Precision: 53& Video & FAUs\\
 &  &  &  Recall: 44& & \\
\hline
\hline
(Yang et al.) \cite{Yang2024} & BiLSTM & Model & Accuracy: 81.10& Video & FAUs \\
 & & & F1-Score: 80.60&  Audio & COVAREP \\
 & & & Precision: 80.20&  Text & BERT \cite{Devlin2018} \\
 & & & Recall: 81&   &  \\
\hline
\hline
(Othmani et al.) \cite{Othmani2022} & CNN & Model & Accuracy*: 87.40& Video & FAUs\\
& & & F1-Score*: 82.30& Audio & VGGish \cite{Simonyan2014} \\
\hline
\hline
(Ceccerali et al.) \cite{Ceccarelli2022} & BiLSTM & Late & F1-Score: 87& Audio & Fisher Vector \cite{Perronnin2010} \\
& & & F1-Score: 70& Video & Fisher Vector \\
& & & Precision: 89& Text & Bag of Words\\
\hline
\hline
(Wei et al.) \cite{Wei2022} & ConvBiLSTM & Model & F1-Score: 0.70& Video & 3D Facial Landmarks\\
& & & Precision: 0.89& & Gaze Tracking \\
& & & & Audio & Log-Mel Spectrogram \\
& & & & Text & Sentence Embeddings \\
\hline
\end{tabular}
\label{beforeandafteraugmentation}
\end{table*}

\subsection{Mel Frequency Cepstral Coefficients in Audio-Based Models}
Speech data has been found to be highly effective, more so than video and text-based data, for diagnosing depression from clinical interview recordings. In fact, multiple mono-modal audio-based models have achieved results comparable to SRIs for diagnosing a variety of psychiatric disorders, such as Generalized Anxiety Disorder (GAD) \cite{Brueckner2024, Zhang2021, Ma2016}. 

Current audio-based diagnosis methods rely on low and high level features extracted from raw audio data. The extracted features are passed as input into deep learning models.

Many representations of audio data have been utilized for this task. (Wang et al.) \cite{Wang2019} evaluated and compared the effectiveness of such representations, finding Mel Frequency Cepstral Coefficients (MFCCs) to be the best performing audial feature for depression diagnosis of the 30 features which were tested in the study.

Derived using the Mel Scale \cite{Pedersen1965}, MFCCs provide a compact representation of the human perception of sound. This allows for the low-level analysis of timbre, pitch, and rhythm. Studies indicate such values are different between depressed and non-depressed patients \cite{Almaghrabi2023}. (Wang et al.) \cite{Wang2023} and (Taguchi et al.) \cite{Taguchi2018} found certain MFCCs follow a consistent pattern that is only present in depressed patients.

Currently, MFCCs are the most used feature across state-of-the-art models to represent the audio mode for this task.

\subsection{Facial Action Units in Video-Based Models}

Research has shown facial expressions and movements are significantly different in depressed patients as opposed to non-depressed ones \cite{Alghowinem2013} \cite{Schneider1990}. Visual-based depression diagnosis models rely on features extracted from patient videos and images.

Facial Action Units (FAUs), a low-level representation of the movements of specific facial muscles, were introduced as part of the Facial Action Coding System \cite{Ekman2019}. FAUs allow us to objectively study facial expressions and decipher non-verbal cues (Figure \ref{fig1}). This makes them highly effective and popular for affective computing tasks \cite{Jiang2022}. In this study, a subset of 20 out of 44 total FAUs are used, as shown in Table \ref{fausavailabledaicwoz}.

\subsection{Multi-Modal Models}

Multi-modal models combine both verbal and non-verbal cues to reach a diagnosis. To our knowledge, the majority of state-of-the-art models for depression diagnosis from clinical interview recordings consider at least audio and video data in making a diagnosis.

\subsubsection{\textbf{Data Fusion Strategies}}

\begin{figure*}[!t]
\centering
\includegraphics[width=7in]{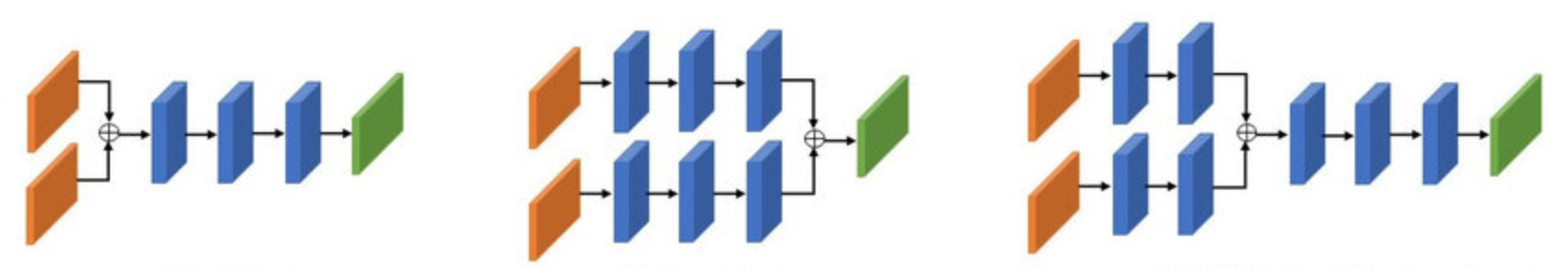}

\caption{Visualization of Early Fusion (left), Late Fusion (middle), and Model-Level Fusion (right) systems \cite{Borges2022}.}
\label{fusion-levels}
\end{figure*}

The way in which the modalities are fused is referred to as the fusion strategy (Figure \ref{fusion-levels}. Two basic fusion strategies which are highly prominent in machine learning models are Early Fusion and Late Fusion \cite{Palowski2023}.

\begin{itemize}
    \item \textbf{Early Fusion:} Representations of different modalities are concatenated before being fed into a neural network. For example, one work combines unprocessed audio and text features before passing them into a Support Vector Machine. However, this method can produce a high-dimensional feature representation, resulting in the 'curse of dimensionality', where the volume of space increases exponentially and training data becomes sparse \cite{Asgari2014}.
    \item \textbf{Late Fusion:} Individual mono-modal models are trained for their respective modalities and their outputs are concatenated right before the final decision. (Samareh et al.) \cite{Samareh2017} employs Late Fusion for MDD diagnosis by evaluating outputs from mono-modal video, audio, and text-based models using Random Forests. When Late Fusion techniques are used, however, deeper complexities and relationships between the different modalities are lost.
\end{itemize}

Model-Level Fusion is a combination of these two strategies. In Model-Level Fusion, individual modalities are processed before concatenation. After concatenation, the data is processed even further to reach an output. This mitigates the curse of dimensionality and issue of uncaptured complexities between modalities. It allows patterns within a single modality and relationships across modalities to be learned effectively. For this reason, it is the most commonly used and best performing fusion method for multi-modal depression diagnosis from clinical interviews.

\section{Data Collection and Preprocessing}

\begin{figure*}[!t]
\centering
\includegraphics[width=6in]{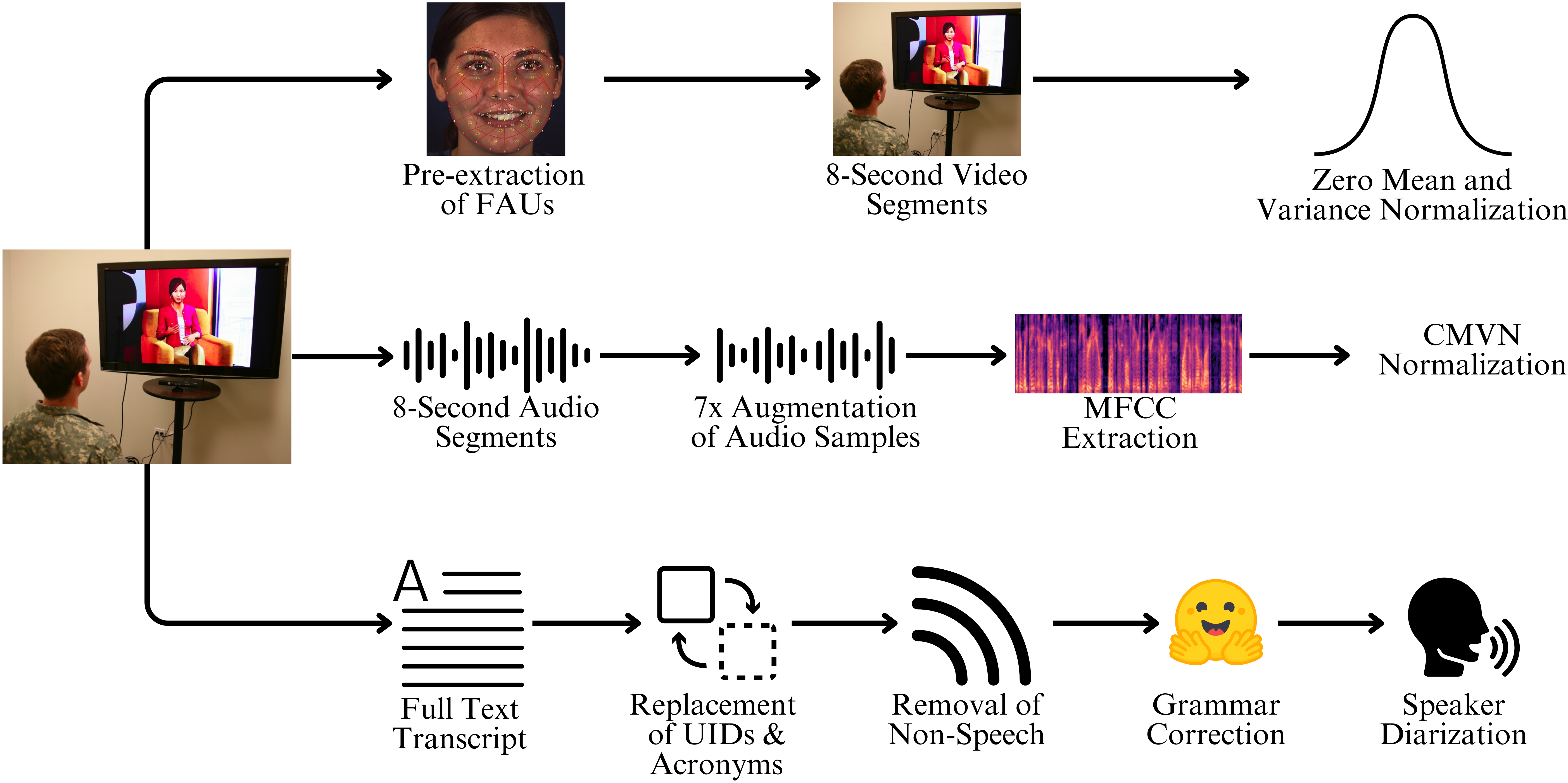}

\caption{Visualization of the pre-processing steps applied to DAIC-WOZ data.}
\label{fig1}
\end{figure*}

\subsection{DAIC-WOZ}
The Distress Analysis Interview Corpus - Wizard of Oz (DAIC-WOZ) contains data collected from 189 clinical interviews ranging between 7-33 minutes \cite{Gratch2014}. Each interview contains a 16kHz .wav recording, a .CSV text transcript, an ID numbered 300 -- 492, and features collected from video data. To keep data de-identified, raw video data is not made available in the DAIC-WOZ. Instead, the dataset provides six visual features extracted in 30 frames per second using the OpenFace pose estimation library \cite{Baltrusaitis2016}. The extracted visual features are as follows: 68 2D Facial Landmarks, 68 3D Facial Landmarks, Histogram of Oriented Gradients, Facial Action Units (FAUs), 4 Gaze Vectors, and Head Pose. FAUs are the only visual feature considered in this study.

In the DAIC-WOZ, depression was assessed using an SRI: Patient Health Questionnare-8 (PHQ-8). The PHQ-8 measures 8 items on a scale of 0 to 3 based on how often problems have been occuring over a two-week period. The final output is a binary label and a score measuring the severity of MDD on a scale between 0-24. A cut-off score of 10 (inclusive) was used for the binary label, where participants scoring higher were considered depressed. These classifications were used to develop, train, and evaluate the proposed architecture.

Access to DAIC-WOZ was granted after signing an End User License Agreement (EULA)\footnote{Zipped DAIC-WOZ files can be found after signing the necessary EULA at \url{https://dcapswoz.ict.usc.edu/.}}.

\subsection{Dataset Errors}
The DAIC-WOZ dataset contains many errors that needed to be resolved before the regular preprocessing steps could be applied. The following errors were resolved:
\begin{enumerate}
    \item All audio files contained interactions between research assistants prior to the interview starting. These pieces of audio were identified and removed.
    \item The interviews with IDs of 373 and 444 contained long interruptions which had to be removed. One such interruption, for example, was the phone of a research assistant ringing.
    \item The interviews with IDs of 451, 458, and 480 were missing the utterances of the therapist. This issue was resolved by manually transcribing the therapist speech.
    \item The interviews with IDs of 318, 321, 341, and 362 contained transcription files which were out of sync with the audio. This was resolved by manually adjusting the timestamps of the affected utterances.
    \item The interview with ID 409 contained a labeling error, where the PHQ-8 score was 10 but the binary label was 0. This was manually resolved.
\end{enumerate}

\subsection{Text Preprocessing}

\begin{table}[!t]
\caption{Transcript Sample From DAIC-WOZ}
\centering
\begin{tabular}{|c|c|c|}
\hline
ID & Speaker & Utterance\\
\hline
0 & Ellie & how\_doingV (so how are you doing today)\\
\hline
1 & Participant & good\\
\hline
2 & Participant & it's been a nice day\\
\hline
3 & Ellie & thats\_good (that's good)\\
\hline
4 & Ellie & where\_originally (where are you from originally)\\
\hline
\end{tabular}
\label{transcriptsample_bad}
\end{table}

\begin{table}[!t]
\caption{Adjusted Transcript Sample}
\centering
\begin{tabular}{|c|c|c|}
\hline
ID & Speaker & Utterance\\
\hline
0 & Therapist & So how are you doing today?\\
\hline
1 & Patient & Good. It's been a nice day.\\
\hline
2 & Therapist & That's good. Where are you from originally?\\
\hline
\end{tabular}
\label{transcriptsample}
\end{table}

The DAIC-WOZ contains transcripts as tab-separated .CSV files with the following columns: speaker, start\_time, stop\_time, and value. The provided utterances contained numerous errors that needed to be fixed.\footnote{Documentation detailing how the DAIC-WOZ can be found online.} Preprocessing was applied using RegEx and NLP techniques to make text suitable to pass into the LLM. Refer to Table \ref{transcriptsample_bad} for a sample of a transcript before pre-processing is applied.

A major issue in the unprocessed transcripts is the presence of unique identifiers within utterances, as seen in Table \ref{transcriptsample_bad}. In the original DAIC-WOZ dataset, all transcriptions of the clinical interviewer were automatically generated for participants whose IDs were greater than 363. In such cases, a unique identifier is followed by the true text in parenthesis. These identifiers are heavily present in Table. RegEx was applied to identify and replace unique identifiers with their true meanings.

In the DAIC-WOZ, acronyms which are pronounced letter by letter are connected by underscores, e.g. 'l\_a' represents 'Los Angeles'. A Python dictionary was generated, where the keys were such acronyms and the values were their extended form. All acronyms were replaced with what they represented.

If speech is cut-off, the full word is followed by what was actually pronounced, enclosed in inequality signs. Non-speech, such as coughing, is enclosed in inequality signs as well. These issues were resolved by simply removing text that followed this pattern using RegEx.

The use of proper grammar in LLM prompting can increase classification accuracy, but the transcripts provided in DAIC-WOZ contain major grammatical errors. To resolve this issue, the 't5-base-grammar-correction' model \cite{Napoles2017} was applied to all text through the Hugging Face platform. The model ensured the proper use of punctuation and capitalization in every utterance. However, the model is limited in its ability to identify interrogative sentences and apply punctuation accordingly. A Support Vector Machine was trained to identify if utterances are interrogative and to apply the appropriate punctuation with an accuracy of 97\%.

The names of diarized speakers were adjusted. 'Ellie' was replaced with 'Therapist' and 'Participant' was replaced with 'Patient'. Using such naming conventions gave the LLM a better understanding of a conversation's context and the speakers' respective roles.

Following all corrections, subsequent utterances by the same speaker were merged together. Refer to Table \ref{transcriptsample} for a sample of a fully adjusted transcript.

The resulting .CSV files were converted into raw text data which could be fed into the LLM. In this raw text format, subsequent utterances were separated by a new-line. Speakers were diarized by preceding utterances with the speaker name, followed by a colon (e.g., 'Therapist: So how are you doing today?').

\subsection{Audio Preprocessing}

Audio preprocessing followed four stages. 
(1) Audio was segmented to only include patient speech. 
(2) Audio data was augmented to increase the number of total samples by a factor of seven. 
(3) MFCCs were extracted using Librosa \cite{McFee2015}. 
(4) The MFCCs were normalized using Cepstral Mean and Variance Normalization.

\subsubsection{\textbf{Audio Segmentation}}
In the proposed model, as well as past research, the audio modality is incorporated by analyzing patient responses to questions. Patient speech was separated from therapist speech by slicing audio according to the timestamps provided in the transcript, which are detailed down to the centisecond. The cropped therapist audio was discarded and not used as input into the final architecture.

The remaining audio data was separated into smaller chunks to make it suitable for input into Bidirectional Long Short Term Memory (BiLSTM) layers. It was sliced into 8-second segments, where any remaining segment under 8 seconds was discarded.

As first done in (Muzammel et al.) \cite{Muzammel2021}, audio data was augmented through pitch shifting and noise injection strategies to prevent over-fitting and data scarcity. In this work, we combined and randomly applied the following augmentation strategies to increase sample size seven-fold:

\begin{itemize}
    \item \textbf{Pitch Shifting}: Pitch was adjusted by 0.5, 2.0, and 2.5 in semitones.
    \item \textbf{Noise Injection}: A NumPy array with randomly assigned values was overlaid onto audio to create noise.
\end{itemize}

\begin{figure}[!t]
\begin{center}
\begin{tikzpicture}[node distance=1.15cm]
\node (start) [startstop] {Speech Waveform};
\node (pro1) [process, below of=start] {Windowing};
\node (pro2) [process, below of=pro1] {Shift it into FFT order};
\node (pro3) [process, below of=pro2] {Find the magnitude of FFT};
\node (pro4) [process, below of=pro3] {Convert the FFT data into filter bank outputs};
\node (pro5) [process, below of=pro4] {Find the log base 10};
\node (pro6) [process, below of=pro5] {Find the cosine transform to reduce dimensionality};
\node (stop) [startstop, below of=pro6] {MFCC Vector};

\draw [arrow] (start) -- (pro1);
\draw [arrow] (pro1) -- (pro2);
\draw [arrow] (pro2) -- (pro3);
\draw [arrow] (pro3) -- (pro4);
\draw [arrow] (pro4) -- (pro5);
\draw [arrow] (pro5) -- (pro6);
\draw [arrow] (pro6) -- (stop);
\newline
\end{tikzpicture}
\end{center}
\caption{Visualization of process used to derive MFCCs. The way in which MFCCs were derived in this work is based off the implementation provided by the Slaney auditory toolbox \cite{Slaney1998}.}
\label{mfccextract}
\end{figure}
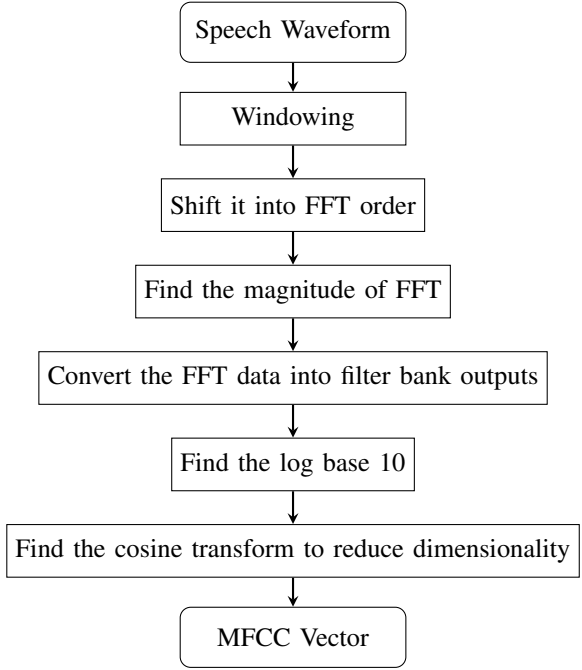

\subsubsection{\textbf{Audio Feature Extraction}}

MFCCs were extracted from raw audio data using the Librosa Python library. 60 MFCCs were extracted from 124ms windows with a 92ms overlap. There were 15,420 total coefficients for every 8-second audio segment. Figure \ref{mfccextract} displays the pipeline used to derive the MFCCs.

\subsubsection{\textbf{MFCC Normalization}}
Rather than normalizing raw audio data, the extracted MFCCs were normalized using Cepstral Mean and Variance Normalization (CMVN). MFCCs are highly sensitive to additive noise and differences in recording conditions. CMVN solves the issue of signal variations due to different recording conditions in MFCCs by normalizing the spectrum to have zero mean and unit variance. Past research has shown applying CMVN to MFCCs greatly improves the performance of speech recognition algorithms \cite{Rehr2015}.

\subsection{Visual Preprocessing}
\begin{table}[!t]
\caption{Facial Action Units Available in DAIC-WOZ.}
\centering
\begin{tabular}{|c||c||c|}
\hline
No. & Feature & Facial Feature Name\\
\hline
1 & AU01\_r & Inner Brow Raiser\\
\hline
2 & AU02\_r & Outer Brow Raiser\\
\hline
3 & AU04\_r & Brow Lowerer\\
\hline
4 & AU05\_r & Upper Lid Raiser\\
\hline
5 & AU06\_r & Cheek Raiser\\
\hline
6 & AU09\_r & Nose Wrinkler\\
\hline
7 & AU10\_r & Upper Lip Raiser\\
\hline
8 & AU12\_r & Lip Corner Puller\\
\hline
9 & AU14\_r & Dimpler\\
\hline
10 & AU15\_r & Lip Corner Depressor\\
\hline
11 & AU17\_r & Chin Raiser\\
\hline
12 & AU20\_r & Lip Stretcher\\
\hline
13 & AU25\_r & Lips Part\\
\hline
14 & AU26\_r & Jaw Drop\\
\hline
15 & AU04\_c & Brow Lowerer\\
\hline
16 & AU12\_c & Lip Corner Puller\\
\hline
17 & AU15\_c & Lip Corner Depressor\\
\hline
18 & AU23\_c & Lip Tightener\\
\hline
19 & AU28\_c & Lip Suck\\
\hline
20 & AU45\_c & Blink\\
\hline
\end{tabular}
\label{fausavailabledaicwoz}
\end{table}

Visual preprocessing followed two stages. (1) Video data was segmented to only include data from when the patient is speaking. (2) Individual columns were normalized for zero mean and unit variance.

\subsubsection{\textbf{Video Segmentation}}
In the proposed architecture, we analyze patient visual data only while they are speaking, not in all parts of the interview. Using the timestamps provided in the transcript, FAU data was cropped to exclude and discard segments where the therapist is speaking, or where nobody is speaking at all. The data was then sliced into 8-second segments in such a way that it aligned with the segmented audio data. Any remaining segment under 8 seconds was discarded. After video segmentation was applied, there were an equal number of video and audio samples for each interview.

\subsubsection{\textbf{FAU Normalization}}
Normalization was applied to continuous, non-discrete FAU values. Namely, numbers 1-14 in Table \ref{fausavailabledaicwoz}. The continuous FAU values were scaled to zero mean and unit variance respectively.
\section{Model Development}

\subsection{Text-Based Model Development}
Two-shot learning was employed with GPT-4 through the OpenAI API to reach a classification given full text transcripts from the DAIC-WOZ.

The model was given the prompt 'Take on the role of an expert in psychiatric diagnosis using the DSM 5. Read the following transcript and determine if the patient has depression.' This prompt is a modified version of that presented in (Galatzer-Levly et al.) \cite{Galatzer-Levy2023}. Given this prompt, GPT-4 was expected to output a binary classification of 'depressed' or 'not depressed' in JSON format. The \textit{function calling} feature provided in the OpenAI API was manipulated to force an output in the desired JSON format in almost all cases.

The examples used as part of few-shot learning consisted of four full-length text transcripts from the DAIC-WOZ dataset. Two of the provided examples had a ground truth of depressed, the other two had a ground truth of not depressed.

\subsection{Tri-Modal Model Development}
Model development followed two stages. (1) The model architecture was optimized with the Hyperband Tuning Algorithm \cite{Li2016} in Keras-Tuner \cite{O'Malley2019}. (2) The model parameters were reset and the optimized architecture was evaluated through Leave-One-Subject-Out Cross-Validation (LOSOCV).

\subsubsection{\textbf{Hyperparameter Tuning}}

\begin{figure*}[!ht]
\begin{center}
\includegraphics[width=6.5in]{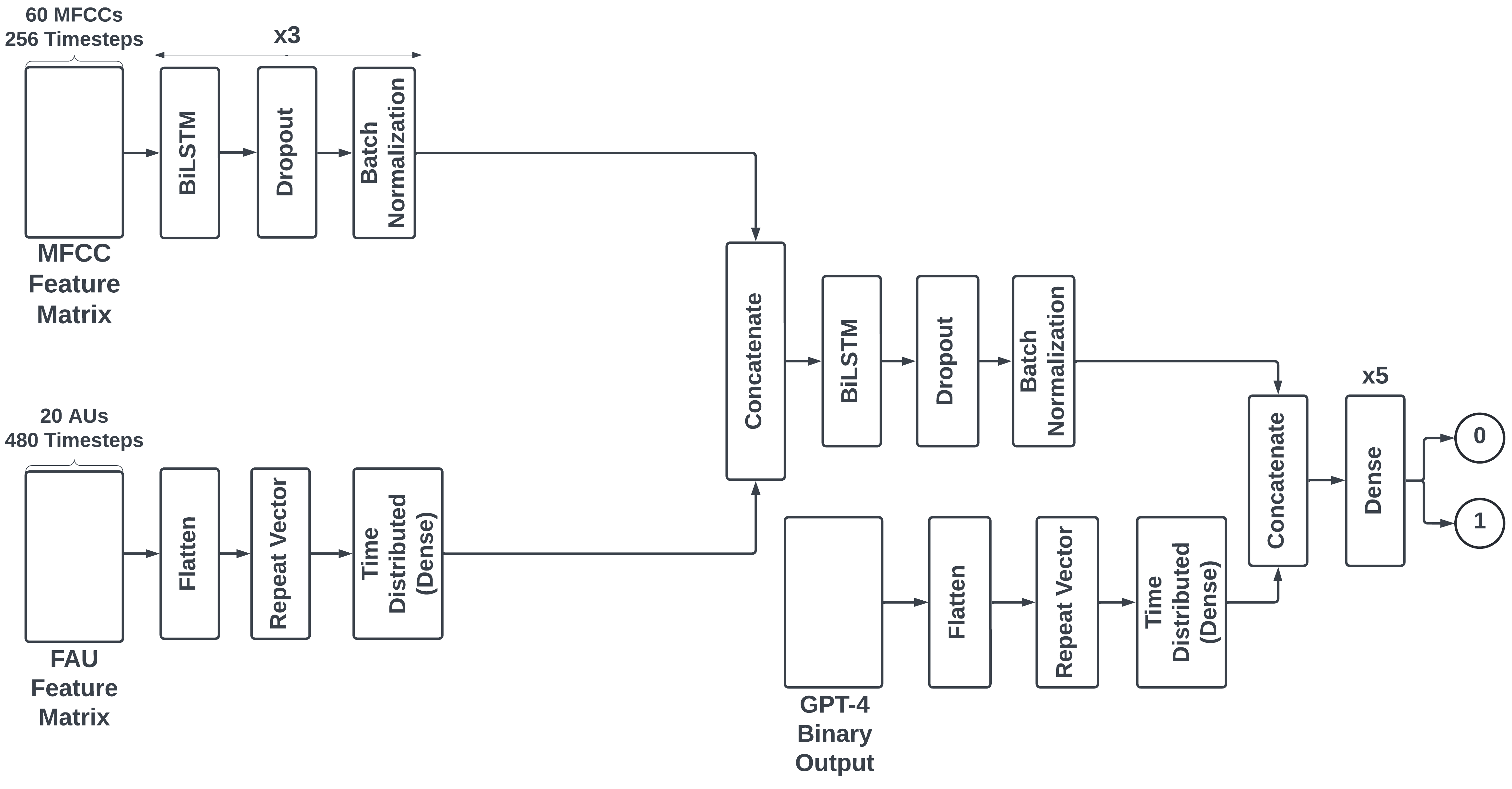}
\end{center}
\caption{Proposed model architecture following Hyperband tuning.}
\label{architecture_final}
\end{figure*}

\begin{table}[!t]
\caption{Hyperparameters optimized through Hyperband}
\centering
\begin{tabular}{|c||c|c|}
\hline
 & Options & Result\\
\hline
BiLSTM 1 Units & 512, 256 & 256\\
\hline
Dropout 1 Rate & 0.1, 0.3, 0.5 & 0.5\\
\hline
BiLSTM 2 Units & 128, 256 & 128\\
\hline
Dropout 2 Rate & 0.1, 0.3, 0.5 & 0.3\\
\hline
BiLSTM 3 Units & 64, 128 & 64\\
\hline
Dropout 3 Rate & 0.1, 0.3, 0.5 & 0.5\\
\hline
Number of Dense Layers & 4, 5, 6 & 4\\
\hline
\end{tabular}
\label{hyperparameters}
\end{table}

\begin{figure}[!t]
\centering
\includegraphics[width=3.5in]{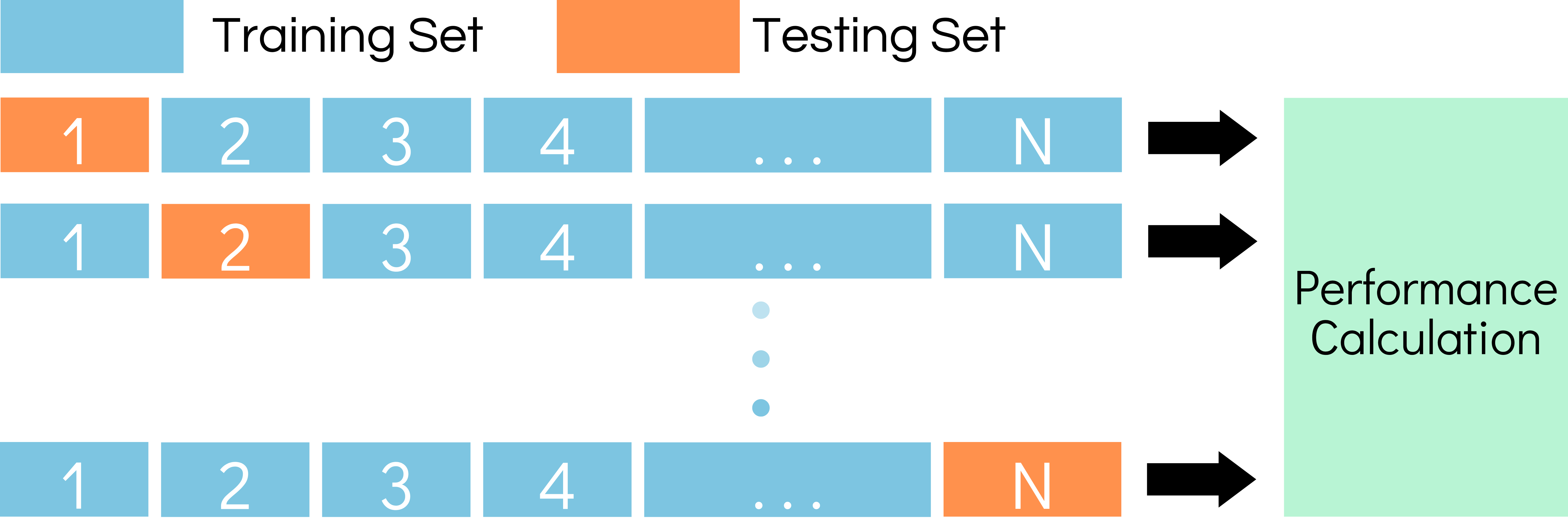}
\caption{Visualization of LOSOCV}
\label{losodiagram}
\end{figure}

The model was trained with the Adam optimizer \cite{Kingma2014} and binary cross entropy as the loss function. To account for the class imbalance of the DAIC-WOZ, the loss function was weighted to give higher value to the depressed class. To achieve this functionality, part of the Keras-Tuner Python library source code was rebuilt.

The tuner was set to focus on validation loss as the only metric to gauge configuration performance. Through each bracket, the number of total epochs run and models tested was reduced by a factor of three. A callback was created to end training and move on to the next configuration if validation loss did not improve over three consecutive epochs. Two iterations of the full Hyperband algorithm were run.

\subsubsection{\textbf{Evaluation}} 
LOSOCV (Figure \ref{losodiagram}) was applied by first resetting the optimized architecture's trainable parameters which had been adjusted during hyperparameter tuning. A unique model was trained for every clinical interview in the dataset (n=180). For each model, all samples originating from one clinical interview were excluded from the training dataset. Each model was tested on the excluded clinical interivew, and all the model results were pooled for the final performance calculation. Augmented samples were not included in testing data.

\subsection{Proposed Architecture}

The final model architecture (Figure \ref{architecture_final}), which has been optimized by the Hyperband Tuning Algorithm, follows 7 steps to reach a classification. 

\begin{enumerate}
    \item The model extracts higher level features from the MFCCs in three consecutive blocks consisting of  BiLSTM, Dropout, and Batch Normalization layers.
    \item The input FAU data follows three processing steps to make it suitable for concatenation with the MFCC tensor: (a) The FAU data is first flattened into a 1D tensor. (b) The Keras \cite{chollet2015keras} \textit{Repeat Vector} layer repeats the flattened data for each timestep present in the processed MFCC tensor, resulting in a 3D tensor. (c) Lastly, the \textit{Time Distributed} operation applies a Dense layer to each temporal slice, where the number of neurons is the number of feature dimensions in the MFCC tensor. These operations result in a higher-level representation of the FAUs with the same dimensions as the processed MFCC data, allowing concatenation in the following step.
    \item The processed FAU and MFCC data are concatenated along the features axis.
    \item The concatenated data is processed by consecutive BiLSTM, Dropout, and Batch Normalization layers.
    \item The same sequence of operations applied in step two is applied to the LLM binary output to prepare it for concatenation with the MFCC-FAU tensor.
    \item The MFCC-FAU and LLM tensors are concatenated.
    \item The combined tensor is processed by four consecutive Leaky ReLU-activated Dense layers and one sigmoid-activated layer. A binary classification is output.
\end{enumerate}

\section{Results}
The proposed architecture was evaluated in two ways:
\begin{enumerate}
    \item The model was evaluated using the DAIC-WOZ AVEC 2016 Challenge train/validation/test split to allow comparability with models from other studies.
    \item The proposed architecture was evaluated through LOSOCV.
\end{enumerate}

\begin{figure*}[!ht]
\centering
\noindent
\renewcommand\arraystretch{1.5}
\setlength\tabcolsep{0pt}
\begin{tabular}{cc}
\makebox[0.5\textwidth]{
\begin{tabular}{c >{\bfseries}r @{\hspace{0.7em}}c @{\hspace{0.4em}}c @{\hspace{0.7em}}l}
  \multirow{10}{*}{\rotatebox{90}{\parbox{1.1cm}{\bfseries\centering Predicted}}} & 
    & \multicolumn{2}{c}{\bfseries Actual} & \\
  & & \bfseries D & \bfseries ND & \bfseries \\
  & D & \MyBox{16}{(34.04)} & \MyBox{1}{(2.13)} \\ [2.4em]
  & ND & \MyBox{3}{(6.38)} & \MyBox{27}{(57.45)} \\
\end{tabular}
}
&
\makebox[0.5\textwidth]{
\begin{tabular}{c >{\bfseries}r @{\hspace{0.7em}}c @{\hspace{0.4em}}c @{\hspace{0.7em}}l}
  \multirow{10}{*}{\rotatebox{90}{\parbox{1.1cm}{\bfseries\centering Predicted}}} & 
    & \multicolumn{2}{c}{\bfseries Actual} & \\
  & & \bfseries D & \bfseries ND & \bfseries total \\
  & D & \MyBox{52}{(27.51)} & \MyBox{13}{(6.88)} & 79.99\\[2.4em]
  & ND & \MyBox{4}{(2.12)} & \MyBox{120}{(63.49)} & 96.77 \\
  & total & 92.85 & 90.22 &
\end{tabular}
}
\end{tabular}
\caption{Confusion matrices of proposed architecture's performance on AVEC 2016 train/validation/test split (left) and LOSOCV (right).}
\end{figure*}

\begin{table*}[!ht]
\caption{Comparison of proposed architecture with baseline binary depression classification models tested on the AVEC 2016 train/validation/test split.}
\centering
\begin{tabular}{|c|c|c|c|c|}
\hline
Paper & Precision & Recall & F1-Score & Accuracy\\
\hline
\hline

(Yang et al.) - Logistic Regression & 53.20 & 55.80 & 54.47 & 54.12\\
(Yang et al.) - Naive Bayes  & 56.70 & 58.40 & 57.54 & 57.02\\
(Yang et al.) - Random Forest & 54.80 & 55.20 & 55.00 & 58.41\\
(Valstar et al.) & 57.90 & 59.60 & 58.74 & 58.41\\
(Qureshi et al.) & 58.00 & 61.00 & 59.46 & 61.11\\
Proposed Model & \textbf{92.86} & \textbf{68.42} & \textbf{78.79} & \textbf{85.11}\\
\hline
\end{tabular}
\label{beforeandafteraugmentation}
\end{table*}

\begin{table*}[!ht]
\caption{Comparison of proposed architecture with state-of-the-art binary depression classification models tested on the AVEC 2016 train/validation/test  split. A dash signifies that the metric was not reported in the paper. (*Evaluation done with LOSOCV)}
\centering
\begin{tabular}{|c|c|c|c|c|c|c|}
\hline
Paper & \multicolumn{2}{|c|}{Precision} & \multicolumn{2}{|c|}{Recall} & F1-Score & Accuracy\\

 & D & ND & D & ND & & \\
\hline
\hline
(Gimeno-Gómez et al.)  & 72 & -- & 72 & -- & 72 & --\\
(Muzammel et al.)  & 53 & 83 & 44 & \textbf{88} & -- & 77.16\\
(Yang et al.)  & 80.20 & -- & 81 & -- & 80.60 & 81.10\\
(Huang et al.)  & \textbf{94} & -- & \textbf{95} & -- & \textbf{94} & \textbf{96}\\
Proposed Model & 92.86 & \textbf{96.43} & 68.42 & 81.82 & 78.79 & 85.11\\
\hline
\hline
(Muzammel et al.)* & \textbf{96} & 95 & 89 & \textbf{98} & \textbf{95.48} & 95.38\\
(Othmani et al.)*  & -- & -- & -- & -- & 82.30 & 87.40\\
(Salekin et al.)*  & -- & -- & -- & -- & 85.44 & \textbf{96.7}\\
Proposed Model* & 80 & 90.23 & \textbf{92.86} & 96.77 & 85.95 & 91.01\\
\hline
\end{tabular}
\end{table*}

The proposed model outperforms all baseline models and multiple state-of-the-art models which were evaluated using no cross-validation split as well as LOSOCV. It performs well on both the depressed and non-depressed classes, with accuracies of 92.9\% and 90.2\% on LOSOCV for the depressed and non-depressed classes, respectively. The large difference in the accuracy between the depressed and non-depressed classes (92.9\% and 81.81\%, respectively) on the AVEC 2016 train/validation/test split can be attributed to the extremely small sample size of just 14 for the depressed class. We believe the accuracy of 92.9\% for the depressed class on the AVEC 2016 Challenge cross-validation split is inflated as a result of this sample size.

\subsection{Computational Complexity}
The speed of the proposed architecture was evaluated on the 12-minute 50-second DAIC-WOZ clinical interview with an ID of 301. The test was conducted using an Intel i7-12650H @ 2.30GHZ processor, NVIDIA RTX 4060 graphics card, and 64GB of RAM. The model took 2.67 seconds to process the interview, including GPT-4 computation time through the OpenAI API.

\section{Deployment}
\begin{figure}[!t]
    \centering
    \includegraphics[width=3.5in]{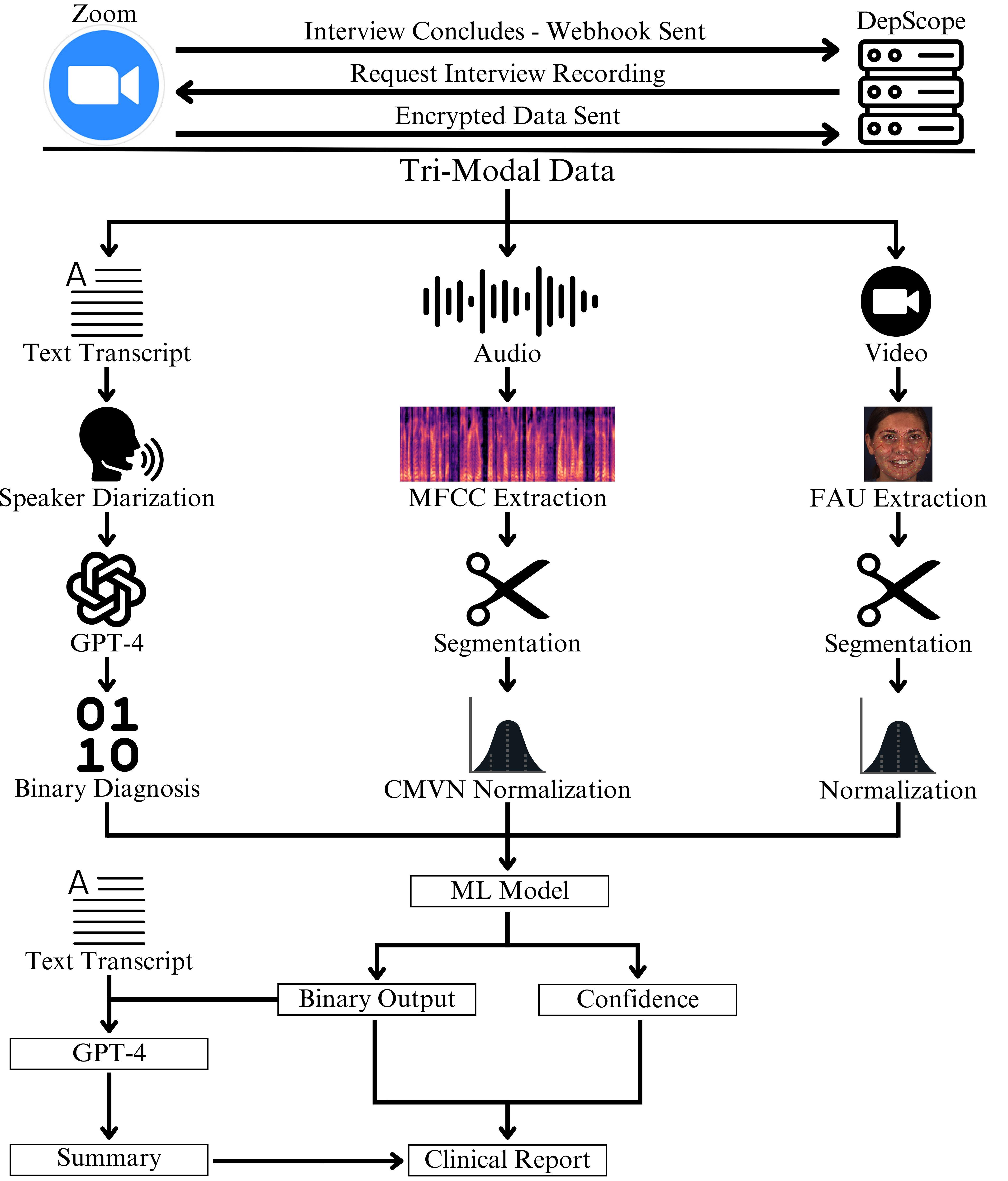}
    \caption{Interview collection and processing pipeline}
    \label{deployment_pipeline}
\end{figure}

\begin{figure}[!t]
    \centering
    \includegraphics[width=2.7in]{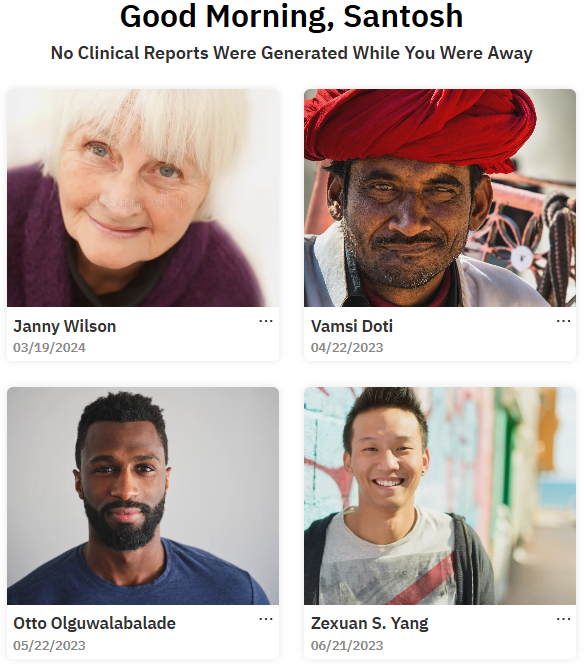}
    \caption{Clinician's dashboard}
    \label{clinician_dashboard}
\end{figure}

\begin{figure*}[!t]
    \centering
    \includegraphics[width=7.5in]{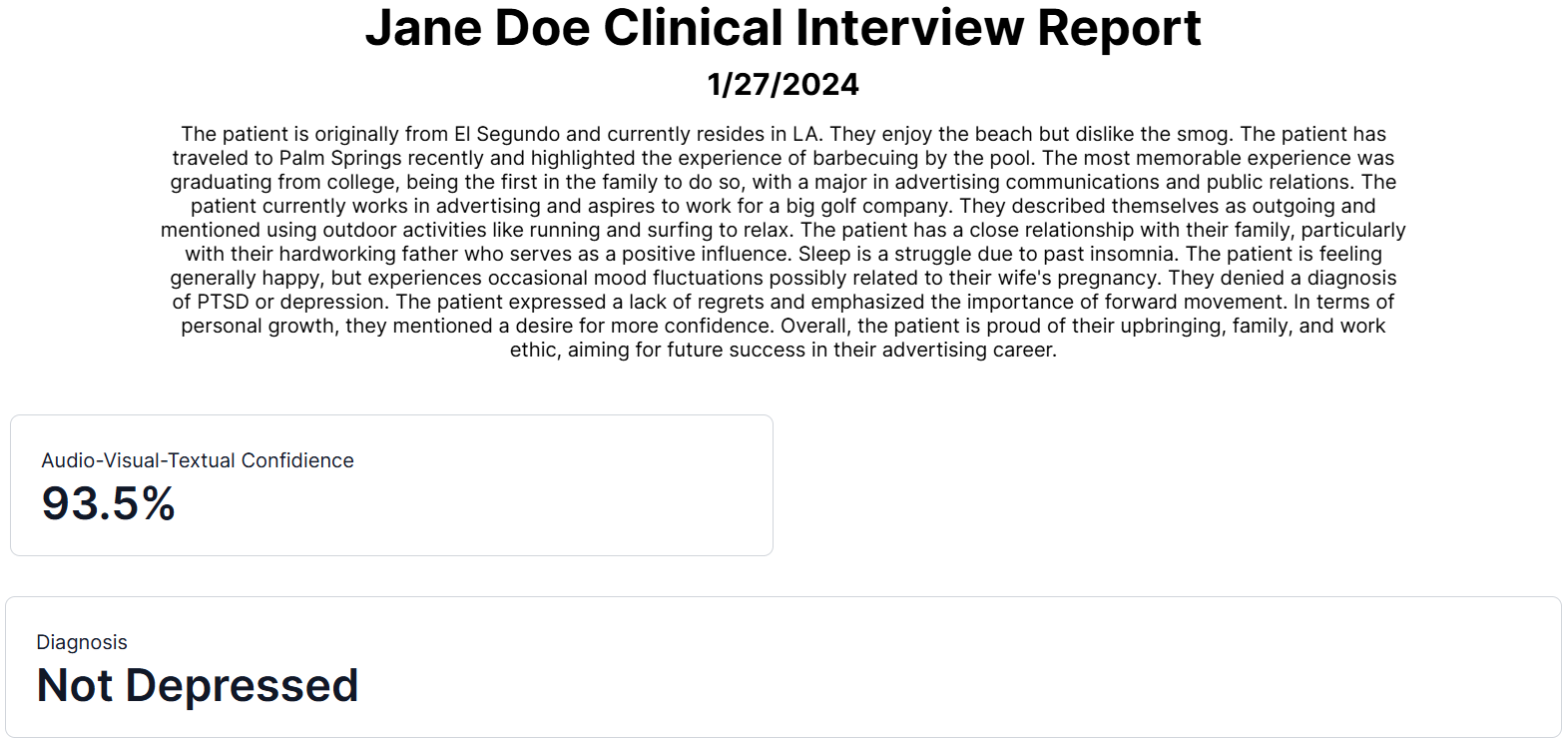}
    \caption{Sample clinical report}
    \label{clinical_report}
\end{figure*}

The proposed architecture was integrated into a locally hosted web application, DepScope, for easy and accessible use of the model by clinicians. This application mimics how such a model may be implemented into real-world scenarios.

Clinicians first connect their Zoom account to the web application and agree to the necessary terms, consenting that their clinical interview recordings may be processed by the DepScope applicatoin. This is done through the Zoom Application Programming Interface (API). From there, any time a clinical interview concludes over Zoom, a webhook \cite{Biehl2017} is sent to the DepScope backend. The recording is then retrieved from the Zoom servers for preprocessing, model inference, and clinical report generation via GPT-4. Clinical reports contain a summary of the interview, detailing all key points and a justification for why the diagnosis was drawn (Figure \ref{clinical_report}). Reports additionally contain the confidence of the model in the classification, which is output by the sigmoid activation function. Reports populate the clinician dashboard (Figure \ref{clinician_dashboard}).

\section{Conclusion}
This paper presents a novel machine learning architecture for diagnosing MDD from clinical interview recordings. We explored the integration of large language models and audiovisual features into a tri-modal architecture for this task. The proposed architecture was optimized through automated hyper-parameter tuning techniques and evaluated using the AVEC 2016 Challenge train/validation/test split and LOSOCV. We drew comparisons between the performance of the proposed architecture and previously developed models, finding that it outperforms all baseline models and multiple state-of-the-art models. The model was integrated into a locally hosted web application to emulate how it may be used in the real-world.

Though the proposed model achieved impressive results, the current architecture faces multiple limitations. The speed of the model suggests that it is not appropriate for real-time applications and may only be effective when integrated with batch-based data processing techniques. Due to the use of large language models in this architecture, it is highly unlikely this can be resolved. The dataset used in this study, the DAIC-WOZ, is small and homogenous. We believe this has negatively impacted the generalizability of the proposed model and may have resulted in metrics which do not reflect the effectiveness of the model in real-world settings. The lack of high-quality data available for this task is an issue which plagues the field as the whole.

In future work, we aim to further optimize the proposed architecture for increased accuracy and make the model easily accessible to use. As new large language models are developed, they will be integrated into the model architecture and evaluated. We plan to integrate existing language models into the architecture as well, such as Med-PaLM 2 \cite{Singhal2023}. The prompt given to the large language model will be improved through iterative prompting techniques. 

\section*{Acknowledgments}
The author would like to extend thanks to the SMU Lyle School of Engineering and the American Psychological Association for their cash award which funded this research.


\nocite{*}

\begin{thebibliography}{10}
\bibliographystyle{IEEEtran}
\providecommand{\url}[1]{#1}
\csname url@samestyle\endcsname
\providecommand{\newblock}{\relax}
\providecommand{\bibinfo}[2]{#2}
\providecommand{\BIBentrySTDinterwordspacing}{\spaceskip=0pt\relax}
\providecommand{\BIBentryALTinterwordstretchfactor}{4}
\providecommand{\BIBentryALTinterwordspacing}{\spaceskip=\fontdimen2\font plus
\BIBentryALTinterwordstretchfactor\fontdimen3\font minus \fontdimen4\font\relax}
\providecommand{\BIBforeignlanguage}[2]{{%
\expandafter\ifx\csname l@#1\endcsname\relax
\typeout{** WARNING: IEEEtran.bst: No hyphenation pattern has been}%
\typeout{** loaded for the language `#1'. Using the pattern for}%
\typeout{** the default language instead.}%
\else
\language=\csname l@#1\endcsname
\fi
#2}}
\providecommand{\BIBdecl}{\relax}
\BIBdecl

\bibitem{Chodavadia2023}
P.~Chodavadia, I.~Teo, D.~Poremski, D.~S.~S. Fung, and E.~A. Finkelstein, ``Prevalence and economic burden of depression and anxiety symptoms among singaporean adults: results from a 2022 web panel,'' \emph{BMC Psychiatry}, vol.~23, p. 104, 2 2023.

\bibitem{Santomauro2021}
D.~F. Santomauro, A.~M.~M. Herrera, J.~Shadid, P.~Zheng, C.~Ashbaugh, D.~M. Pigott, C.~Abbafati, C.~Adolph, J.~O. Amlag, A.~Y. Aravkin, B.~L. Bang-Jensen, G.~J. Bertolacci, S.~S. Bloom, R.~Castellano, E.~Castro, S.~Chakrabarti, J.~Chattopadhyay, R.~M. Cogen, J.~K. Collins, X.~Dai, W.~J. Dangel, C.~Dapper, A.~Deen, M.~Erickson, S.~B. Ewald, A.~D. Flaxman, J.~J. Frostad, N.~Fullman, J.~R. Giles, A.~Z. Giref, G.~Guo, J.~He, M.~Helak, E.~N. Hulland, B.~Idrisov, A.~Lindstrom, E.~Linebarger, P.~A. Lotufo, R.~Lozano, B.~Magistro, D.~C. Malta, J.~C. Månsson, F.~Marinho, A.~H. Mokdad, L.~Monasta, P.~Naik, S.~Nomura, J.~K. O'Halloran, S.~M. Ostroff, M.~Pasovic, L.~Penberthy, R.~C.~R. Jr, G.~Reinke, A.~L.~P. Ribeiro, A.~Sholokhov, R.~J.~D. Sorensen, E.~Varavikova, A.~T. Vo, R.~Walcott, S.~Watson, C.~S. Wiysonge, B.~Zigler, S.~I. Hay, T.~Vos, C.~J.~L. Murray, H.~A. Whiteford, and A.~J. Ferrari, ``Global prevalence and burden of depressive and anxiety disorders in 204 countries and territories in 2020 due to the covid-19
  pandemic,'' \emph{The Lancet}, vol. 398, pp. 1700--1712, 11 2021.

\bibitem{Guidi2011}
J.~Guidi, G.~A. Fava, P.~Bech, and E.~Paykel, ``The clinical interview for depression: A comprehensive review of studies and clinimetric properties,'' \emph{Psychotherapy and Psychosomatics}, vol.~80, pp. 10--27, 2011.

\bibitem{Radloff1977}
L.~S. Radloff, ``The ces-d scale,'' \emph{Applied Psychological Measurement}, vol.~1, pp. 385--401, 6 1977.

\bibitem{Kroenke2001}
K.~Kroenke, R.~L. Spitzer, and J.~B.~W. Williams, ``The phq-9,'' \emph{Journal of General Internal Medicine}, vol.~16, pp. 606--613, 9 2001.

\bibitem{Allsopp2019}
Allsopp, K., Read, J., Corcoran, R. \& Kinderman, P. Heterogeneity in psychiatric diagnostic classification. {\em Psychiatry Res.}. \textbf{279} pp. 15--22, 9 2019.

\bibitem{De_Silva2017}
Silva, P. How to improve psychiatric services: a perspective from critical psychiatry. {\em Br. J. Hosp. Med. (Lond.)}. \textbf{78}, 503--507, 9 2017.

\bibitem{Fuchs2010}
Fuchs, T. Subjectivity and intersubjectivity in psychiatric diagnosis. {\em Psychopathology}. \textbf{43}, 268--274, 5 2010.

\bibitem{Althubaiti2016}
A.~Althubaiti, ``Information bias in health research: definition, pitfalls, and adjustment methods,'' \emph{Journal of Multidisciplinary Healthcare}, p. 211, 5 2016.

\bibitem{Latkin2017}
C.~A. Latkin, C.~Edwards, M.~A. Davey-Rothwell, and K.~E. Tobin, ``The relationship between social desirability bias and self-reports of health, substance use, and social network factors among urban substance users in baltimore, maryland,'' \emph{Addictive Behaviors}, vol.~73, pp. 133--136, 10 2017.

\bibitem{Mendel2011}
R.~Mendel, E.~Traut-Mattausch, E.~Jonas, S.~Leucht, J.~M. Kane, K.~Maino, W.~Kissling, and J.~Hamann, ``Confirmation bias: why psychiatrists stick to wrong preliminary diagnoses,'' \emph{Psychological Medicine}, vol.~41, pp. 2651--2659, 12 2011.

\bibitem{Ayano2021}
G.~Ayano, S.~Demelash, Z.~yohannes, K.~Haile, M.~Tulu, D.~Assefa, A.~Tesfaye, K.~Haile, M.~Solomon, A.~Chaka, and L.~Tsegay, ``Misdiagnosis, detection rate, and associated factors of severe psychiatric disorders in specialized psychiatry centers in ethiopia,'' \emph{Annals of General Psychiatry}, vol.~20, p.~10, 2 2021.

\bibitem{Muzammel2021}
M.~Muzammel, H.~Salam, and A.~Othmani, ``End-to-end multimodal clinical depression recognition using deep neural networks: A comparative analysis,'' \emph{Computer Methods and Programs in Biomedicine}, vol. 211, p. 106433, 11 2021.

\bibitem{Ma2016}
X.~Ma, H.~Yang, Q.~Chen, D.~Huang, and Y.~Wang, ``Depaudionet.''\hskip 1em plus 0.5em minus 0.4em\relax ACM, 10 2016, pp. 35--42.

\bibitem{Brueckner2024}
R.~Brueckner, N.~Kwon, V.~Subramanian, N.~Blaylock, and H.~O’Connell, \emph{Audio-Based Detection of Anxiety and Depression via Vocal Biomarkers}, 2024, pp. 124--141.

\bibitem{Zhang2021}
P.~Zhang, M.~Wu, H.~Dinkel, and K.~Yu, ``Depa.''\hskip 1em plus 0.5em minus 0.4em\relax ACM, 10 2021, pp. 135--143.

\bibitem{Wang2019}
J.~Wang, L.~Zhang, T.~Liu, W.~Pan, B.~Hu, and T.~Zhu, ``Acoustic differences between healthy and depressed people: a cross-situation study,'' \emph{BMC Psychiatry}, vol.~19, p. 300, 12 2019.

\bibitem{Tu2019}
C.-H. Tu, C.-Y. Yang, and J.~Y. jen Hsu, ``Idennet: Identity-aware facial action unit detection.''\hskip 1em plus 0.5em minus 0.4em\relax IEEE, 5 2019, pp. 1--8.

\bibitem{Gimeno}
D.~Gimeno-Gómez, A.-M. Bucur, A.~Cosma, C.-D. Martínez-Hinarejos, and P.~Rosso, \emph{Reading Between the Frames: Multi-modal Depression Detection in Videos from Non-verbal Cues}, 2024, pp. 191--209.

\bibitem{Degottex2014}
Degottex, G., Kane, J., Drugman, T., Raitio, T., and Scherer, S. ``COVAREP — A collaborative voice analysis repository for speech technologies.`` {\em 2014 IEEE International Conference On Acoustics, Speech And Signal Processing (ICASSP)}, 5 2014.

\bibitem{Sataloff2017}
Sataloff, R. Vocal Health and pedagogy: Science, assessment, and treatment, Third Edition. (Plural Publishing), 9 2017.

\bibitem{Yang2024}
S.~Yang, L.~Cui, L.~Wang, T.~Wang, and J.~You, ``Enhancing multimodal depression diagnosis through representation learning and knowledge transfer,'' \emph{Heliyon}, vol.~10, p. e25959, 2 2024.

\bibitem{Devlin2018}
Devlin, J., Chang, M., Lee, K., and Toutanova, K. BERT: Pre-training of deep bidirectional Transformers for language understanding. (arXiv,2018)

\bibitem{Othmani2022}
A.~Othmani, A.-O. Zeghina, and M.~Muzammel, ``A model of normality inspired deep learning framework for depression relapse prediction using audiovisual data,'' \emph{Computer Methods and Programs in Biomedicine}, vol. 226, p. 107132, 11 2022.

\bibitem{Simonyan2014}
Simonyan, K., and Zisserman, A. ``Very deep convolutional networks for large-scale image recognition.`` \emph{arXiv}, 2014.

\bibitem{Ceccarelli2022}
F.~Ceccarelli, M.~Mahmoud, ``Multimodal temporal machine learning for Bipolar Disorder and Depression Recognition,`` \emph{Pattern Anal. Appl.}. \textbf{25}, 493-504, 8 2022. 

\bibitem{Perronnin2010}
Perronnin, F., Sánchez, J. \& Mensink, T. Improving the fisher kernel for large-scale image classification. {\em Computer Vision – ECCV 2010}. pp. 143-156 (2010)

\bibitem{Wei2022}P.~Wei, K.~Peng, A.~Roitberg, K.~Yang, J.~Zhang, and R.~Stiefelhagen, ``Multi-modal Depression Estimation based on Sub-attentional Fusion.,`` \emph{arXiv}, 2022

\bibitem{Pedersen1965}Pedersen, P. The Mel scale. {\em J. Music Theory}. \textbf{9}, 295 (1965)

\bibitem{Almaghrabi2023}
S.~A. Almaghrabi, S.~R. Clark, and M.~Baumert, ``Bio-acoustic features of depression: A review,'' \emph{Biomedical Signal Processing and Control}, vol.~85, p. 105020, 8 2023.

\bibitem{Wang2023}Wang, Y., Liang, L., Zhang, Z., Xu, X., Liu, R., Fang, H., Zhang, R., Wei, Y., Liu, Z., Zhu, R., Zhang, X. \& Wang, F. Fast and accurate assessment of depression based on voice acoustic features: a cross-sectional and longitudinal study. {\em Front. Psychiatry}. \textbf{14} pp. 1195276 (2023,6)

\bibitem{Taguchi2018}
T.~Taguchi, H.~Tachikawa, K.~Nemoto, M.~Suzuki, T.~Nagano, R.~Tachibana, M.~Nishimura, and T.~Arai, ``Major depressive disorder discrimination using vocal acoustic features,'' \emph{Journal of Affective Disorders}, vol. 225, pp. 214--220, 1 2018.

\bibitem{Alghowinem2013}
S.~Alghowinem, R.~Goecke, M.~Wagner, G.~Parkerx, and M.~Breakspear, ``Head pose and movement analysis as an indicator of depression.''\hskip 1em plus 0.5em minus 0.4em\relax IEEE, 9 2013, pp. 283--288.

\bibitem{Schneider1990}
F.~Schneider, H.~Heimann, W.~Himer, D.~Huss, R.~Mattes, and B.~Adam, ``Computer-based analysis of facial action in schizophrenic and depressed patients,'' \emph{European Archives of Psychiatry and Clinical Neuroscience}, vol. 240, pp. 67--76, 11 1990.

\bibitem{Ekman2019}
Ekman, P. \& Friesen, W. Facial Action Coding System. (American Psychological Association (APA),2019,1), Title of the publication associated with this dataset: PsycTESTS Dataset

\bibitem{Jiang2022}
Jiang, Z., Luskus, M., Seyedi, S., Griner, E., Rad, A., Clifford, G., Boazak, M. \& Cotes, R. Utilizing computer vision for facial behavior analysis in schizophrenia studies: A systematic review. {\em PLoS One}. \textbf{17}, e0266828 (2022,4)

\bibitem{Borges2022}
Borges, P., Peynot, T., Liang, S., Arain, B., Wildie, M., Minareci, M., Lichman, S., Samvedi, G., Sa, I., Hudson, N., Milford, M., Moghadam, P. \& Corke, P. A survey on terrain traversability analysis for autonomous ground vehicles: Methods, sensors, and challenges. {\em Field Robotics}. \textbf{2}, 1567-1627 (2022,3)

\bibitem{Palowski2023}
M.~Pawłowski, A.~Wróblewska, and S.~Sysko-Romańczuk, ``Effective techniques for multimodal data fusion: A comparative analysis,'' \emph{Sensors}, vol.~23, p. 2381, 2 2023.

\bibitem{Asgari2014}
M.~Asgari, I.~Shafran, and L.~B. Sheeber, ``Inferring clinical depression from speech and spoken utterances.''\hskip 1em plus 0.5em minus 0.4em\relax IEEE, 9 2014, pp. 1--5.

\bibitem{Samareh2017}
A.~Samareh, Y.~Jin, Z.~Wang, X.~Chang, and S.~Huang, ``Predicting depression severity by multi-modal feature engineering and fusion,'' 11 2017.

\bibitem{Gratch2014}
\BIBentryALTinterwordspacing
J.~Gratch, R.~Artstein, G.~Lucas, G.~Stratou, S.~Scherer, A.~Nazarian, R.~Wood, J.~Boberg, D.~DeVault, S.~Marsella, D.~Traum, S.~Rizzo, and L.-P. Morency, ``The distress analysis interview corpus of human and computer interviews,'' N.~Calzolari, K.~Choukri, T.~Declerck, H.~Loftsson, B.~Maegaard, J.~Mariani, A.~Moreno, J.~Odijk, and S.~Piperidis, Eds.\hskip 1em plus 0.5em minus 0.4em\relax European Language Resources Association (ELRA), 5 2014, pp. 3123--3128. [Online]. Available: \url{http://www.lrec-conf.org/proceedings/lrec2014/pdf/508_Paper.pdf}
\BIBentrySTDinterwordspacing

\bibitem{Baltrusaitis2016}
Baltrusaitis, T., Robinson, P. \& Morency, L. OpenFace: An open source facial behavior analysis toolkit. {\em 2016 IEEE Winter Conference On Applications Of Computer Vision (WACV)}. (2016,3)

\bibitem{Napoles2017}Napoles, C., Sakaguchi, K. \& Tetreault, J. JFLEG: A fluency corpus and benchmark for grammatical error correction. (arXiv,2017)

\bibitem{McFee2015}
B.~McFee, C.~Raffel, D.~Liang, D.~Ellis, M.~McVicar, E.~Battenberg, and O.~Nieto, ``librosa: Audio and music signal analysis in python,'' 2015, pp. 18--24.

\bibitem{Slaney1998}
M.~Slaney, ``Auditory toolbox,'' \emph{Interval Research Corporation, Tech. Rep}, vol.~10, p. 1194, 1998.

\bibitem{Rehr2015}
R.~Rehr and T.~Gerkmann, ``Cepstral noise subtraction for robust automatic speech recognition.''\hskip 1em plus 0.5em minus 0.4em\relax IEEE, 4 2015, pp. 375--378.

\bibitem{Galatzer-Levy2023}
I.~R. Galatzer-Levy, D.~McDuff, V.~Natarajan, A.~Karthikesalingam, and M.~Malgaroli, ``The capability of large language models to measure psychiatric functioning,'' 8 2023.

\bibitem{Li2016}
L.~Li, K.~Jamieson, G.~DeSalvo, A.~Rostamizadeh, and A.~Talwalkar, ``Hyperband: A novel bandit-based approach to hyperparameter optimization,'' 3 2016.

\bibitem{O'Malley2019}
T.~O'Malley, E.~Bursztein, J.~Long, F.~Chollet, H.~Jin, L.~Invernizzi \emph{et~al.}, ``Kerastuner,'' 2019.

\bibitem{Kingma2014}
D.~P. Kingma and J.~Ba, ``Adam: A method for stochastic optimization,'' 12 2014.

\bibitem{Valstar2016}
M.~Valstar, J.~Gratch, B.~Schuller, F.~Ringeval, D.~Lalanne, M.~T. Torres, S.~Scherer, G.~Stratou, R.~Cowie, and M.~Pantic, ``Avec 2016.''\hskip 1em plus 0.5em minus 0.4em\relax ACM, 10 2016, pp. 3--10.

\bibitem{Qureshi2019}
S.~A. Qureshi, S.~Saha, M.~Hasanuzzaman, and G.~Dias, ``Multitask representation learning for multimodal estimation of depression level,'' \emph{IEEE Intelligent Systems}, vol.~34, pp. 45--52, 9 2019.

\bibitem{chollet2015keras}
Chollet, F. \& Others Keras. (GitHub,2015), https://github.com/fchollet/keras

\bibitem{Huang2024}
X.~Huang, F.~Wang, Y.~Gao, Y.~Liao, W.~Zhang, L.~Zhang, and Z.~Xu, ``Depression recognition using voice-based pre-training model,'' \emph{Scientific Reports}, vol.~14, p. 12734, 6 2024.

\bibitem{Salekin2018}
A.~Salekin, J.~W. Eberle, J.~J. Glenn, B.~A. Teachman, and J.~A. Stankovic, ``A weakly supervised learning framework for detecting social anxiety and depression,'' \emph{Proceedings of the ACM on Interactive, Mobile, Wearable and Ubiquitous Technologies}, vol.~2, pp. 1--26, 7 2018.

\bibitem{Biehl2017}
Biehl, M. Webhooks. (Createspace Independent Publishing Platform,2017,12)

\bibitem{Singhal2023}
Singhal, K., Tu, T., Gottweis, J., Sayres, R., Wulczyn, E., Hou, L., Clark, K., Pfohl, S., Cole-Lewis, H., Neal, D., Schaekermann, M., Wang, A., Amin, M., Lachgar, S., Mansfield, P., Prakash, S., Green, B., Dominowska, E., Arcas, B., Tomasev, N., Liu, Y., Wong, R., Semturs, C., Mahdavi, S., Barral, J., Webster, D., Corrado, G., Matias, Y., Azizi, S., Karthikesalingam, A. \& Natarajan, V. Towards expert-level medical question answering with large language models. (arXiv,2023)

\end{thebibliography}
\end{document}